\documentclass[letterpaper]{article} 
\usepackage[preprint]{aaai2027}  
\usepackage[hyphens]{url}  
\usepackage{graphicx} 
\usepackage{natbib}  
\usepackage{caption} 
\usepackage{algorithm}
\usepackage{algorithmic}
\usepackage{booktabs}
\usepackage{amsmath}
\usepackage{amssymb}

\usepackage{newfloat}
\usepackage{listings}
\DeclareCaptionStyle{ruled}{labelfont=normalfont,labelsep=colon,strut=off} 
\floatstyle{ruled}
\newfloat{listing}{tb}{lst}{}
\floatname{listing}{Listing}

\usepackage{booktabs}

\title{TIER-MoE: Trust-Informed Expert Routing via Conditional Modality Risk for Multimodal Fusion in Biomedical Classification}

\author{
    Yu Chang\equalcontrib\textsuperscript{\rm 1},
    Anzhe Cheng\equalcontrib\textsuperscript{\rm 1},
    Chenwei Wu\equalcontrib\textsuperscript{\rm 2},\\
    Zhuoran Wang\textsuperscript{\rm 3},
    Jiahao Chen\textsuperscript{\rm 4},
    Tamoghna Chattopadhyay\textsuperscript{\rm 1},\\
    Sophia I. Thomopoulos\textsuperscript{\rm 1},
    Paul M. Thompson\textsuperscript{\rm 1},
    Liyue Shen\textsuperscript{\rm 2},\\
    Paul Bogdan\corresponding\textsuperscript{\rm 1}
}
\affiliations{
    \textsuperscript{\rm 1}University of Southern California\\
    \textsuperscript{\rm 2}University of Michigan\\
    \textsuperscript{\rm 3}The University of British Columbia\\
    \textsuperscript{\rm 4}University of Toronto\\
    pbogdan@usc.edu
}

\begin{document}

\maketitle

\begin{abstract}

The promise of multimodal fusion lies in combining complementary sources of evidence, yet more evidence does not always yield a better prediction. Recent multimodal models have advanced fusion through richer cross-modal interaction and sample-adaptive fusion. However, the influence assigned to a modality during fusion does not reveal whether that source is unreliable, redundant, or poorly matched to a specialized expert. To address this limitation, we introduce
\textbf{TIER-MoE}, a risk-guided subspace mixture-of-experts model
that defines sample-specific modality reliability as the prediction
loss its unimodal predictor is expected to incur. This risk is learned
from out-of-fold predictions generated by models that were not trained
on the corresponding sample.
TIER-MoE combines the estimated risk with expert-specific subspace compatibility for sparse modality--expert routing, while an
always-active shared path preserves multimodal complementarity. We evaluate TIER-MoE on four public multimodal biomedical datasets spanning Alzheimer's disease status, skin-lesion malignancy, and retinal classification. Results demonstrate its superiority over state-of-the-art methods in predictive performance and probability calibration, with consistent improvements in Macro-F1 and Brier score and strong zero-shot generalization to an external cohort.

\end{abstract}



\section{Introduction}
\label{sec:introduction}
Multimodal biomedical classification is widely used for disease diagnosis, prognosis, and patient stratification~\cite{doan2026bridging}. 
As many diseases affect several biological systems simultaneously, a single measurement often provides only a partial view of the underlying condition. Multimodal biomedical prediction addresses this limitation by combining observations of anatomy, tissue microstructure, physiology, and clinical state \cite{acosta2022multimodal,stahlschmidt2022multimodal}. For example, T1-weighted (T1w) MRI captures macroscopic brain anatomy, diffusion tensor imaging (DTI) reflects white-matter microstructure~\cite{lorio2016neurobiological,tae2018clinical}, and structured clinical metadata provide complementary information about patient state and disease presentation~\cite{acosta2022multimodal}. In principle, these modalities should improve prediction by providing complementary evidence. However, in practice, their quality and predictive value vary across subjects~\cite{han2022multimodal,zhang2023provable}, and multimodal models can even perform worse than their strongest unimodal components in some cases~\cite{huang2022modality}. 
The main difficulty is therefore not simply how to combine multiple representations, but how to determine which evidence is reliable for each subject without discarding information that remains useful through cross-modal interaction.

\begin{figure}[t]
    \centering
    \includegraphics[width=\linewidth]{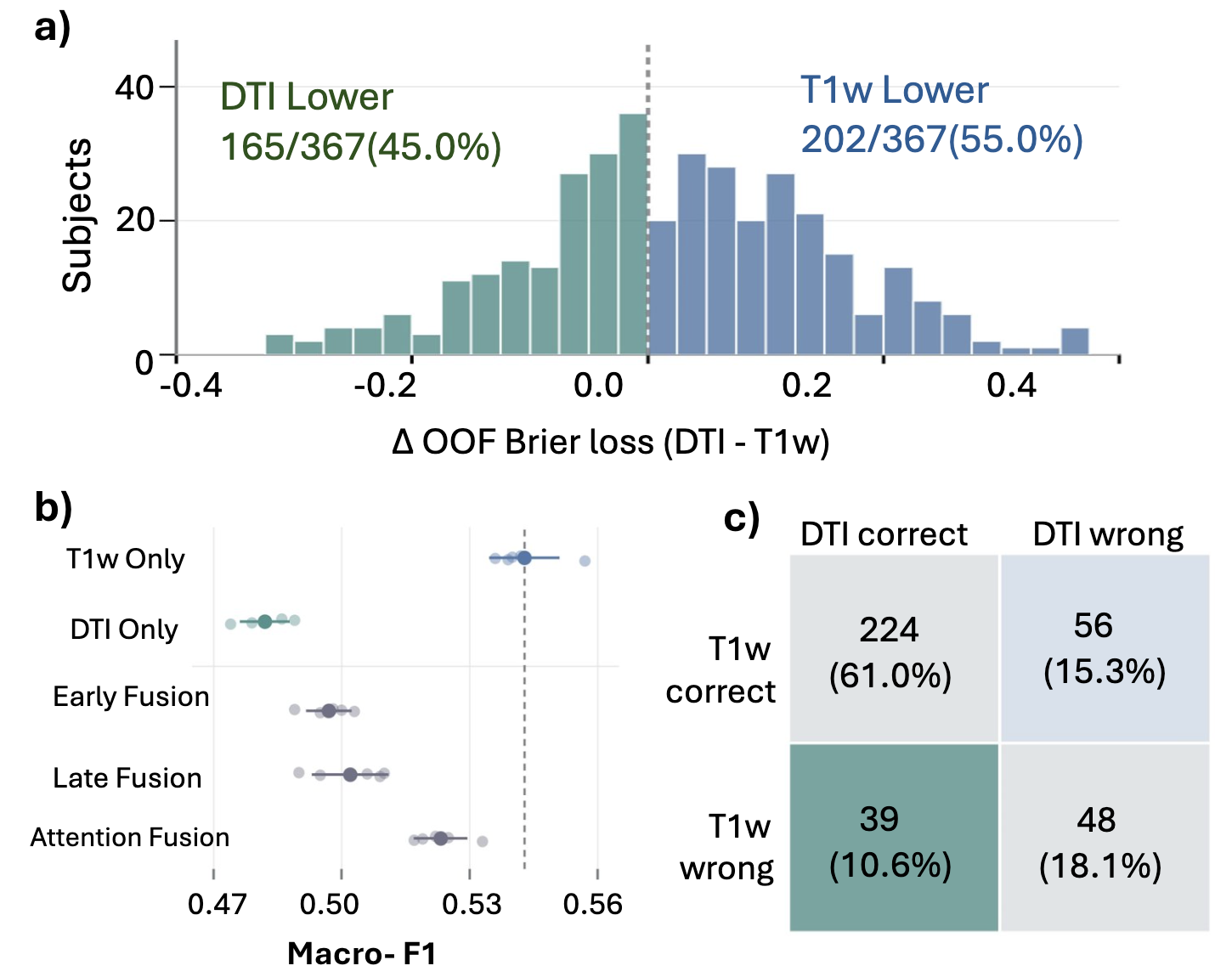}
    \caption{Motivation for subject-adaptive fusion.
    (a) The modality with lower subject-level OOF Brier loss varies across subjects.
    (b) Standard fusion baselines remain below T1w across repeated runs.
    (c) DTI is uniquely correct for 39 of 367 subjects, demonstrating nonredundant evidence.}
    \label{fig:motivation}
\end{figure}

Existing fusion methods do not fully resolve this reliability--complementarity trade-off~\cite{han2022multimodal,neverova2016moddrop,wang2020training}. Jointly trained fusion architectures include gated models such as GMU~\cite{arevalo2017gated} and attention-based models such as MBT~\cite{nagrani2021attention}. Under joint optimization, these models can over-rely on the globally dominant
modality, inducing \emph{modality competition} that underutilizes
complementary evidence from weaker modalities and may leave fusion below
the best unimodal predictor~\cite{huang2022modality,wang2020training}. Adaptive fusion methods adjust modality influence per sample using
evidential uncertainty in TMC~\cite{han2023trusted},
quality-aware weighting in QMF~\cite{zhang2023provable},
collaborative confidence in PDF~\cite{cao2024predictive}, or
probabilistic credibility in Cred-MF~\cite{sidheekh2025credibility}. 
Although these methods adapt modality influence across samples, their scores represent uncertainty, quality, confidence, or credibility rather than directly predicting the loss that a modality-specific classifier is expected to incur on the current sample. More importantly, reliability is usually used directly to control fusion: once a modality is judged less reliable, its contribution is reduced correspondingly. This treatment conflates two distinct cases: a modality that is
unreliable for the current subject and one that is weak in isolation
but complementary in fusion. Our ADNI analysis illustrates this
distinction. As shown in Fig.~\ref{fig:motivation}, the more reliable modality,
as indicated by lower subject-level OOF Brier loss, varies across
subjects (a), yet conventional T1w--DTI fusion remains below the T1w
predictor across repeated runs (b). Although DTI is weaker in aggregate, it uniquely produces the correct prediction for 39 of 367 subjects, demonstrating nonredundant complementary evidence (c). Thus, global modality strength does not describe subject-level reliability, and subject-level reliability alone does not measure the complementary value of a modality.

\begin{figure*}[!ht]
    \centering
    \includegraphics[width=\textwidth]{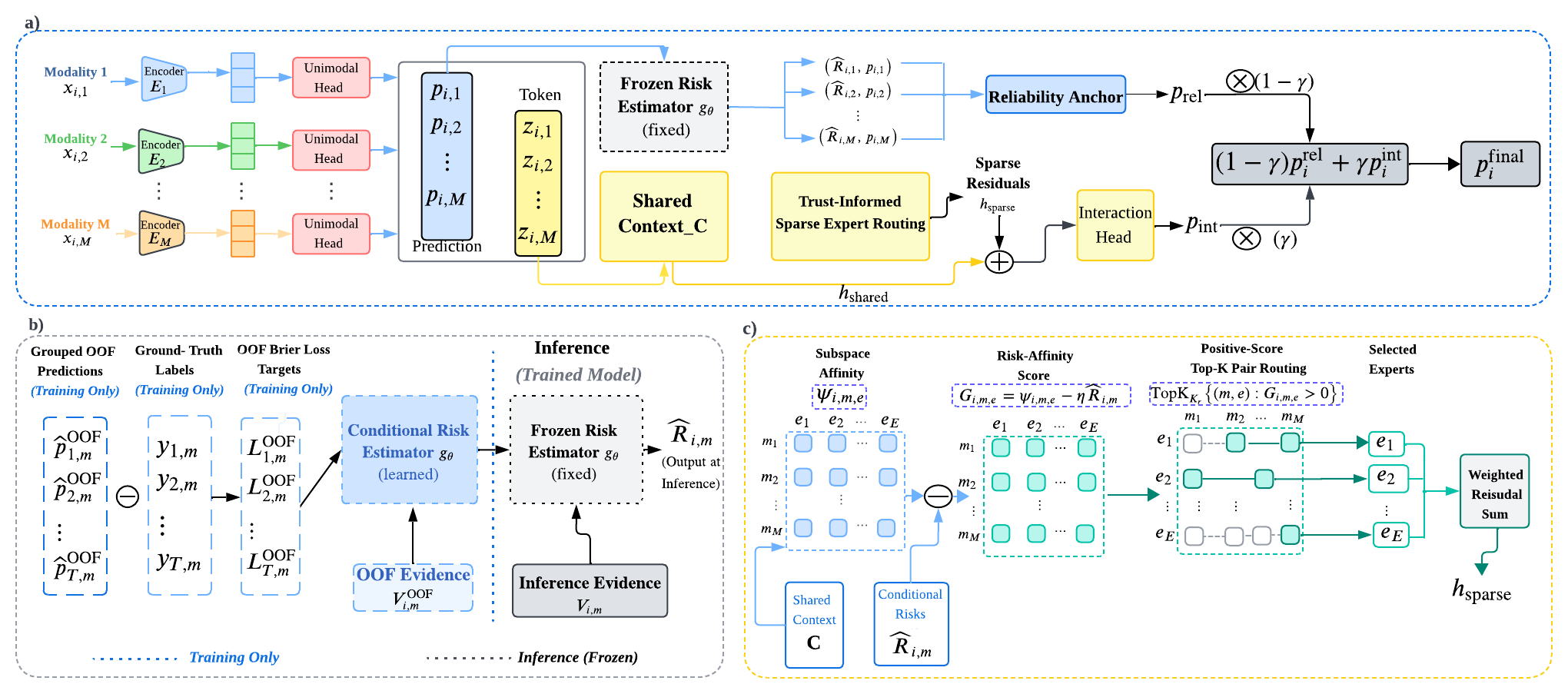}
    \caption{Overview of \textsc{TIER-MoE}.
(a) Calibrated unimodal predictions form a risk-weighted reliability
branch, while shared context and risk-guided sparse experts form an
interaction branch; their predictions are mixed by $\gamma$.
(b) The risk estimator is trained on grouped OOF evidence and realized
Brier-loss targets, then frozen.
(c) Positive risk-adjusted subspace-affinity scores select Top-$K$
modality--expert routes whose weighted residuals augment the shared
path.}
    \label{fig:architecture}
\end{figure*}

Beyond modality reliability, biomedical image classification is further complicated by heterogeneity in disease-relevant patterns. Patients sharing the same diagnostic label can exhibit distinct atrophy
trajectories, while structural MRI and DTI emphasize different disease-relevant regions and provide complementary diagnostic information~\cite{poulakis2022multicohort,bron2017multiparametric}. This heterogeneity suggests that structural alterations may dominate for some subjects, whereas diffusion-related changes or their interaction
with structural features may be more informative for others. A single fusion function is unlikely to represent all such patterns equally well. Mixture-of-experts (MoE) models provide conditional computation through specialized experts that are selectively activated for each input \cite{shazeer2017outrageously,riquelme2021scaling}. Recent multimodal MoE models instantiate expert specialization in different ways: M4oE combines modality-specific and shared modality--task experts
~\cite{wu2025dynamic}, Flex-MoE routes different observed modality
combinations~\cite{yun2024flexmoe}, and I$^2$MoE models distinct
cross-modal interactions with specialized experts
~\cite{xin2025i2moe}. ERMoE further aligns expert selection with
learned representation subspaces~\cite{cheng2026ermoe}. However, these
methods do not jointly couple task-grounded, subject-specific modality
risk with expert compatibility. Thus, modality reliability,
cross-modal complementarity, and expert specialization remain
uncoupled in existing routing formulations.

To address these limitations, we propose the \emph{Trust-Informed Expert Routing via Conditional Modality Risk} (\textsc{TIER-MoE}), a hierarchical framework that connects subject-specific modality reliability estimation with expert specialization for multimodal fusion. 
\textsc{TIER-MoE} first estimates the conditional risk of each modality, defined as the expected prediction loss of its unimodal predictor for the current subject. For each subject, the risk target is the loss of an out-of-fold unimodal prediction generated by a model trained on the remaining
subjects. These estimates form a reliability anchor that preserves dependable unimodal evidence without acting as the sole fusion mechanism. In parallel, \textsc{TIER-MoE} learns low-rank expert subspaces and measures their affinity with each modality representation. A joint risk--affinity criterion then selects sparse modality--expert routes, assigning reliable evidence to experts whose subspaces are suitable for processing it. A shared interaction path retains information across all available modalities, while routed expert residuals capture specialized subject-dependent interactions. This design assigns distinct roles to three quantities that prior
methods either conflate or model separately: the task risk of each
modality-specific predictor, its complementary value during fusion,
and its compatibility with a specialized expert.

The primary contributions of this work are summarized as follows:
\begin{itemize}
    \item We formulate subject-specific modality reliability as the conditional expected prediction loss of a unimodal predictor and learn it from subject-level out-of-fold predictions. This definition provides a common, task-grounded reliability measure that can be evaluated directly against held-out prediction errors.
    \item We introduce \textsc{TIER-MoE}, which combines conditional modality risk with expert-subspace affinity for subject-dependent modality--expert routing. Its reliability anchor and shared interaction path preserve reliable unimodal predictions while retaining complementary evidence from modalities that are weaker in isolation.
    \item We evaluate \textsc{TIER-MoE} across four biomedical cohorts and three classification tasks under standard prediction, modality degradation, missingness, cross-modal conflict, and external cohort shift, demonstrating consistent improvements in prediction, calibration, robustness, and harmful fusion reduction.
\end{itemize}

 

\section{Method}
\label{sec:method}

\subsection{Network Architecture}
\label{sec:architecture}

Figure~\ref{fig:architecture} illustrates the proposed
\textbf{TIER-MoE}, short for \emph{Trust-Informed Expert Routing via Conditional Modality Risk}.
The framework assigns a distinct mechanism to each quantity identified in the introduction: conditional risk estimates the expected loss of each modality-specific predictor, an always-active shared path maintains cross-modal interaction independently of sparse routing, and
expert-subspace affinity measures representation-level compatibility.
The reliability branch forms a risk-weighted unimodal prediction,
whereas the interaction branch adds sparsely routed expert residuals
to the shared representation. Their predictions are combined using a
validation-locked coefficient.


\subsection{Cross-Fitted Conditional Modality Risk}
\label{sec:risk}

We define sample-specific modality reliability through
\emph{conditional modality risk}, the expected prediction loss of the corresponding
unimodal predictor:
\begin{equation}
R_m(V_i)
=
\mathbb{E}\!\left[
L_{i,m}\mid V_i
\right].
\label{eq:conditional_risk}
\end{equation}
where
$V_i=\{(v_{i,m},a_{i,m})\}_{m=1}^{M}$
collects inference-time evidence across modalities.
The local vector $v_{i,m}$ contains the calibrated predictive
distribution, predictive uncertainty, technical quality indicators,
cross-modal disagreement, and fold-comparable representation
summaries, while $a_{i,m}$ indicates modality availability.

\paragraph{Label-honest risk targets.}
The conditional expectation in Eq.~\eqref{eq:conditional_risk} is not directly
observed. We construct its realized training targets from out-of-fold (OOF)
unimodal predictions. For multiclass classification, we use the normalized
Brier loss:
\begin{equation}
L_{i,m}^{\mathrm{OOF}}
=
\frac{1}{2}
\left\|
p_{i,m}^{\mathrm{OOF}}-e_{Y_i}
\right\|_2^2,
\label{eq:oof_loss}
\end{equation}
where $e_{Y_i}$ is the one-hot label vector and
$p_{i,m}^{\mathrm{OOF}}$ is the calibrated prediction of a unimodal model
trained without subject $i$. The normalization gives
$L_{i,m}^{\mathrm{OOF}}\in[0,1]$.

We obtain these targets through grouped subject-level cross-fitting. For each
fold, the complete supervised unimodal branch is trained on the remaining
subjects. Its probability calibrator is fitted using only data within that
fold's training partition. The calibrated model is then evaluated on the
held-out subjects to obtain $p_{i,m}^{\mathrm{OOF}}$. Thus, the label of
subject $i$ is used to evaluate its realized loss but never to fit the
predictor or calibrator that produces its OOF prediction.

\paragraph{Risk estimator.}
A shared mask-aware estimator embeds each modality's local evidence and
conditions its risk on the pooled context of all available modalities:
\begin{equation}
\begin{aligned}
u_{i,m}
&=
\phi_{\theta}([v_{i,m},e_m]),\\
r_{i,m}^{\mathrm{raw}}
&=
\sigma\!\left(
h_{\theta}\!\left(
\left[
u_{i,m},
\frac{
\sum_{j=1}^{M}a_{i,j}\xi_{\theta}(u_{i,j})
}{
\sum_{j=1}^{M}a_{i,j}
}
\right]
\right)
\right),
\end{aligned}
\label{eq:raw_risk_prediction}
\end{equation}
where $e_m$ is a learned modality embedding and
$\phi_{\theta}$, $\xi_{\theta}$, and $h_{\theta}$ are shared
nonlinear mappings.

The estimator is supervised by the realized OOF Brier losses:
\begin{equation}
\mathcal{L}_{\mathrm{risk}}
=
\frac{
\displaystyle
\sum_{i=1}^{N}
\sum_{m=1}^{M}
a_{i,m}
\left(
r_{i,m}^{\mathrm{raw}}
-
L_{i,m}^{\mathrm{OOF}}
\right)^2
}{
\displaystyle
\sum_{i=1}^{N}
\sum_{m=1}^{M}
a_{i,m}
}.
\label{eq:risk_objective}
\end{equation}
Squared-loss regression targets the conditional expectation
in Eq.~(1). We train the risk estimator on all subject-level
OOF evidence--loss pairs and freeze it before optimizing the
interaction branch. For interaction training, we store the
OOF-derived risk estimates
\[
R^{\mathrm{OOF}}_{i,m}
=
g_{\theta}(V^{\mathrm{OOF}}_i,m),
\]
where $g_{\theta}$ denotes the estimator in Eq.~(3).

\paragraph{Risk recalibration.}
After refitting the unimodal predictors on the complete training
partition and fitting their probability calibrators on the validation
set, the frozen risk estimator produces $r^{\mathrm{raw}}_{i,m}$ from
the final-model evidence. To correct the OOF-to-refit shift, we fit a
low-capacity monotone map on the validation set:
\begin{equation}
\begin{aligned}
\widehat{R}_{i,m}
&=
C_m\!\left(r^{\mathrm{raw}}_{i,m}\right),
\\
C_m(r)
&=
\sigma\!\left(
a\,\operatorname{logit}(r)+b_m
\right),
\qquad a>0.
\end{aligned}
\label{eq:risk_recalibration}
\end{equation}
The target is the realized normalized Brier loss of the
probability-calibrated refit unimodal predictor. The map is frozen
for test and external evaluation.

\paragraph{Reliability anchor.}
The calibrated risks define an absolute reliability score and a
relative allocation over the available modalities:
\begin{equation}
\begin{aligned}
t_{i,m}
&=
a_{i,m}\exp(-\beta\widehat R_{i,m}),\\
\alpha_{i,m}
&=
\frac{\pi_m t_{i,m}}
{\sum_{j=1}^{M}\pi_j t_{i,j}},\\
p_i^{\mathrm{rel}}
&=
\sum_{m=1}^{M}\alpha_{i,m}p_{i,m},
\end{aligned}
\label{eq:reliability_anchor}
\end{equation}
where $\beta>0$ controls allocation concentration and
$\pi_m>0$ is a fixed modality prior, chosen uniformly in all
experiments. Here, $t_{i,m}$ is an absolute reliability score,
whereas $\alpha_{i,m}$ is a relative allocation among available
modalities.

\subsection{Risk-Guided Subspace Expert Routing}
\label{sec:routing}

Let $\tilde{z}_{i,m}$ denote modality $m$ projected into the common
latent space and augmented with its modality embedding, and let
$z_{\mathrm{CLS}}$ be a learnable shared token. Availability-masked
self-attention produces contextualized modality and shared tokens:
\[
\begin{aligned}
&\bigl[
z_{i,\mathrm{CLS}},z_{i,1},\ldots,z_{i,M}
\bigr]
\\
&\quad =
\operatorname{MaskedAttn}\!\left(
\bigl[
z_{\mathrm{CLS}},
\tilde{z}_{i,1},\ldots,\tilde{z}_{i,M}
\bigr];
\mathbf{a}_i
\right).
\end{aligned}
\]
Let $A_{i,mj}$ denote the resulting attention weight from modality
$m$ to an available modality $j$. We define
\[
\begin{aligned}
c_{i,m}
&=
\sum_{j:\,a_{i,j}=1}
A_{i,mj}z_{i,j},
\\
h_i^{\mathrm{shared}}
&=
H_{\mathrm{sh}}\!\left(z_{i,\mathrm{CLS}}\right).
\end{aligned}
\]
The shared representation is computed from all available modalities
independently of risk weighting and sparse route selection, maintaining
an always-active path for cross-modal interaction.

In parallel with the shared path, TIER-MoE maintains a bank of $E$
sparse residual experts. For routing, each expert
$e\in\{1,\ldots,E\}$ is associated with a low-rank subspace
$B_e\in\mathbb{R}^{d\times r}$, where $r<d$. Its columns are encouraged to be orthonormal during training. Inspired by representation-aligned expert routing, ERMoE \cite{cheng2026ermoe}, we define modality--expert compatibility as

\begin{equation}
\psi_{i,m,e}
=
\bar{z}_{i,m}^{\top}
B_e B_e^{\top}
\bar{c}_{i,m},
\label{eq:subspace_affinity}
\end{equation}
where $\bar{z}_{i,m}$ and $\bar{c}_{i,m}$ are $\ell_2$-normalized before
projection. The compatibility score approaches zero when either the modality token or its
context has little support in the expert subspace. It therefore expresses
expert compatibility rather than modality reliability. During interaction-branch training, Eq.~\eqref{eq:risk_affinity}
uses the stored OOF-derived estimate $R^{\mathrm{OOF}}_{i,m}$.
Validation, test, and external evaluation use the recalibrated
final-model risk $\widehat R_{i,m}$ from
Eq.~\eqref{eq:risk_recalibration}.

Risk and compatibility are combined at the modality--expert-pair level:
\begin{equation}
G_{i,m,e}
=
\psi_{i,m,e}
-
\eta\widehat R_{i,m}.
\label{eq:risk_affinity}
\end{equation}
where $\eta \geq 0$ controls the risk penalty. We consider only
available modality--expert pairs with positive scores and retain at
most $K_r$ routes:
\begin{equation}
\begin{aligned}
\mathcal{C}_i
&=
\left\{(m,e)\mid a_{i,m}=1,\;G_{i,m,e}>0\right\},\\
\mathcal{S}_i
&=
\operatorname{TopK}_{K_r}\!\left(\mathcal{C}_i;G_i\right).
\end{aligned}
\label{eq:route_selection}
\end{equation}

The compatibility term rewards expert-subspace alignment, whereas
conditional risk penalizes evidence likely to incur prediction loss. Thus, a moderately risky modality may remain routable when strongly aligned
with an expert, while low risk alone does not force assignment to an
unsuitable expert. Cross-modal interaction remains available through
the shared path.

\subsection{Interaction Prediction and Training}
\label{sec:interaction}

For a nonempty selected set $\mathcal{S}_i$, the risk-adjusted routing scores
are normalized over the activated modality--expert pairs:
\begin{equation}
\rho_{i,m,e}
=
\frac{\exp(G_{i,m,e}/\tau)}
{\displaystyle\sum_{(j,e')\in\mathcal{S}_i}
\exp(G_{i,j,e'}/\tau)},
\qquad
(m,e)\in\mathcal{S}_i,
\label{eq:routing_weights}
\end{equation}
where $\tau>0$ controls the concentration of the routing distribution. A
smaller $\tau$ places more weight on the highest-scoring routes, whereas a
larger $\tau$ produces a softer allocation among the selected pairs.

Each selected expert processes its assigned modality representation through a
low-rank residual transformation:
\begin{equation}
r_e(z)
=
U_e\phi\!\left(D_eB_e^{\top}z\right),
\label{eq:expert_residual}
\end{equation}
where $B_e$ projects the input into expert $e$'s routing subspace,
$D_e$ is a learned diagonal transformation, $U_e$ maps the transformed
features back to the shared representation space, and $\phi$ is a nonlinear
activation.

The selected expert outputs first form a sparse, sample-specific correction.
This correction is then added to the always-active shared representation, and
the resulting interaction representation is mapped to a predictive
distribution:
\begin{equation}
\begin{aligned}
h_i^{\mathrm{sparse}}
&=
\sum_{(m,e)\in\mathcal{S}_i}
\rho_{i,m,e}\,
r_e(z_{i,m}),\\
h_i^{\mathrm{int}}
&=
h_i^{\mathrm{shared}}
+
h_i^{\mathrm{sparse}},\\
p_i^{\mathrm{int}}
&=
\operatorname{softmax}\!\left(
H_{\mathrm{int}}(h_i^{\mathrm{int}})
\right).
\end{aligned}
\label{eq:interaction_prediction}
\end{equation}
Here, $h_i^{\mathrm{sparse}}$ contains only the selected expert
residuals. Because $h_i^{\mathrm{shared}}$ is computed independently
of $\mathcal S_i$, an empty route set yields
$h_i^{\mathrm{int}}=h_i^{\mathrm{shared}}$. Null routing therefore
suppresses sparse expert computation without removing the shared
cross-modal path, and only selected experts are evaluated.

To balance reliability-guided unimodal evidence with cross-modal
interaction, we combine the two branch predictions:
\begin{equation}
p_i^{\mathrm{final}}
=
(1-\gamma)p_i^{\mathrm{rel}}
+
\gamma p_i^{\mathrm{int}},
\label{eq:final_prediction}
\end{equation}
where $\gamma\in[0,1]$ is selected on the validation set and fixed for test and external evaluation.

\paragraph{Interaction objective.}
Following representation-aligned expert parameterization
\cite{cheng2026ermoe}, we optimize the interaction branch with the
task objective and two complementary subspace regularizers:
\begin{equation}
\mathcal{L}_{\mathrm{int}}
=
\mathcal{L}_{\mathrm{task}}
+
\lambda_{\mathrm{orth}}\mathcal{L}_{\mathrm{orth}}
+
\lambda_{\mathrm{sep}}\mathcal{L}_{\mathrm{sep}}.
\label{eq:interaction_objective}
\end{equation}
Here, $\mathcal{L}_{\mathrm{orth}}$ promotes semi-orthogonal input
and output bases within each expert, while
$\mathcal{L}_{\mathrm{sep}}$ encourages distinct routing subspaces
across experts.

\paragraph{Training protocol.}
We first generate subject-level OOF unimodal predictions, losses, and
risk evidence within the training partition. The risk estimator is
trained on all OOF evidence--loss pairs and then frozen. We next refit
the unimodal predictors on the complete training partition, fit their
probability calibrators and the final risk map $C_m$ on validation,
and freeze them. With the final encoders and risk estimator fixed, the
interaction branch is trained using the stored OOF-derived risk
estimates. Finally, both prediction branches are frozen and $\gamma$
is selected on validation for locked test and external evaluation.

\section{Experiments}

\begin{table*}[t]
\centering
\begin{small}
\setlength{\tabcolsep}{1mm}

\begin{tabular*}{\textwidth}{
    @{\extracolsep{\fill}}lcccccccccccc@{}
}
\toprule
& \multicolumn{4}{c}{\textbf{ADNI}}
& \multicolumn{4}{c}{\textbf{PAD-UFES-20}}
& \multicolumn{4}{c}{\textbf{FPRM}} \\
\cmidrule(lr){2-5}
\cmidrule(lr){6-9}
\cmidrule(l){10-13}

\textbf{Method}
& \textbf{AUROC} $\uparrow$
& \textbf{F1} $\uparrow$
& \textbf{Brier} $\downarrow$
& \textbf{ECE} $\downarrow$
& \textbf{AUROC} $\uparrow$
& \textbf{F1} $\uparrow$
& \textbf{Brier} $\downarrow$
& \textbf{ECE} $\downarrow$
& \textbf{AUROC} $\uparrow$
& \textbf{F1} $\uparrow$
& \textbf{Brier} $\downarrow$
& \textbf{ECE} $\downarrow$ \\
\midrule

Modality 1 only
& $0.764$ & $0.523$ & $0.286$ & $0.104$
& $0.845$ & $0.752$ & $0.158$ & $0.071$
& $0.746$ & $0.508$ & $0.294$ & $0.123$ \\

Modality 2 only
& $0.731$ & $0.482$ & $0.309$ & $0.121$
& $0.782$ & $0.681$ & $0.204$ & $0.101$
& $0.701$ & $0.461$ & $0.328$ & $0.148$ \\

\midrule

DAFT
& $0.798$ & $0.562$ & $0.257$ & $0.082$
& $0.874$ & $0.781$ & $0.143$ & $0.061$
& $0.775$ & $0.546$ & $0.278$ & $0.109$ \\

GMU
& $0.811$ & $0.578$ & $0.315$ & $0.118$
& $0.885$ & $0.795$ & $0.136$ & $0.056$
& $0.789$ & $0.559$ & $0.267$ & $0.101$ \\

\midrule

Flex-MoE
& $0.829$ & $0.596$ & $0.234$ & $0.069$
& $0.897$ & $0.811$ & $0.129$ & $0.049$
& $0.804$ & $0.574$ & $0.252$ & $0.086$ \\

I$^{2}$MoE
& $0.841$ & $\underline{0.614}$ & $0.223$ & $0.061$
& $0.906$ & $0.823$ & $0.123$ & $0.044$
& $0.817$ & $0.596$ & $0.241$ & $0.078$ \\

\midrule

QMF
& $0.835$ & $0.602$ & $0.222$ & $0.050$
& $\underline{0.913}$ & $0.838$ & $0.116$ & $0.036$
& $\underline{0.838}$ & $0.615$ & $0.229$ & $0.060$ \\

PDF
& $0.838$ & $0.591$ & $\underline{0.214}$
& $\mathbf{0.038}$
& $0.910$ & $\underline{0.841}$
& $\underline{0.108}$ & $\underline{0.031}$
& $0.831$ & $\underline{0.622}$
& $\underline{0.215}$ & $\underline{0.052}$ \\

\midrule

OOF MetaGate
& $0.842$ & $0.607$ & $0.219$ & $0.056$
& $0.905$ & $0.826$ & $0.119$ & $0.041$
& $0.825$ & $0.602$ & $0.236$ & $0.071$ \\

Affinity-MoE
& $\mathbf{0.849}$ & $0.610$ & $0.226$ & $0.059$
& $0.911$ & $0.832$ & $0.111$ & $0.039$
& $0.834$ & $0.619$ & $0.219$ & $0.057$ \\

\midrule

\textbf{TIER-MoE}
& $\underline{0.846}$ & $\mathbf{0.648}$
& $\mathbf{0.189}$ & $\underline{0.041}$
& $\mathbf{0.938}$ & $\mathbf{0.884}$
& $\mathbf{0.094}$ & $\mathbf{0.023}$
& $\mathbf{0.855}$ & $\mathbf{0.657}$
& $\mathbf{0.198}$ & $\mathbf{0.041}$ \\

\bottomrule
\end{tabular*}
\end{small}

\caption{Overall predictive and probabilistic performance across the three in-domain biomedical datasets. AUROC is macro one-vs-rest for ADNI and FPRM and binary for PAD-UFES-20; F1 denotes Macro-F1. In all of the following tables, the \textbf{bold} denotes the best result and \underline{underline} the second best result.}
\label{tab:overall_performance}
\end{table*}

We evaluate \textsc{TIER-MoE} in terms of
(1) in-domain prediction and probability quality,
(2) conditional-risk validity,
(3) robustness to unreliable evidence,
(4) zero-shot cross-cohort transfer, and
(5) component necessity.

\subsection{Experimental Setup}
\label{sec:experiment_setup}

\paragraph{Datasets and splits.}
We evaluate three in-domain modality regimes: T1w MRI and DTI for
three-class AD/MCI/CN classification on ADNI~\cite{petersen2010adni};
lesion images and clinical metadata for binary malignancy
classification on PAD-UFES-20~\cite{pacheco2020padufes20}; and fundus
photographs and retinal blood-flow videos for three-class retinal
classification on FPRM~\cite{zhang2024fprm}. We use fixed
group-disjoint train/validation/test splits of 5:1:1,
4.5:1:1, and 4:1:1, respectively. All observations from the
same subject or patient remain in one split, and PAD-UFES-20
predictions are aggregated at the lesion level. OASIS-3%
~\cite{lamontagne2019oasis3} is reserved for frozen
ADNI-to-OASIS-3 evaluation without adaptation.

\paragraph{Evaluation protocol.}
Results are averaged over three prespecified seeds using identical
grouped splits, and no seed is selected or replaced based on test
performance. All model selection and post-hoc calibration are
performed on the validation set and frozen before test evaluation.
For each method and seed, scalar temperature scaling is fitted by
minimizing validation negative log-likelihood.

We report Macro-F1 as the primary predictive metric, accompanied by Area Under the Receiver Operating Characteristic Curve (AUROC),
Brier score~\cite{brier1950verification}, and Expected Calibration Error (ECE)~\cite{guo2017calibration}. Balanced accuracy
is additionally reported for cross-cohort evaluation. Modality-risk
metrics are computed on the held-out ADNI test set using the frozen
final unimodal predictors. ERCE is the mean absolute gap between predicted risk and realized unimodal Brier loss over ten equal-frequency bins; High-Loss AUROC defines high loss using the 80th percentile of the pooled validation-set unimodal Brier losses; and Correct-Modality Risk Accuracy (CMRA) is the fraction of ADNI test subjects with exactly one correct unimodal prediction for which the correct modality receives strictly lower predicted risk than the incorrect modality.

For robustness evaluation, all methods are trained only on clean data.
The same fixed severity grid of controlled T1w and DTI degradation is
applied to the held-out ADNI test inputs for every method. Retention AUC integrates clean-normalized Macro-F1 over degradation severity.
Missing F1 is the unweighted mean of Macro-F1 under the T1w-missing
and DTI-missing conditions. Conflict F1 is evaluated on subjects whose
two unimodal argmax predictions disagree. Harm Rate is the proportion
of incorrect fused predictions among subjects for which at least one
unimodal predictor is correct. Clean Harm is evaluated on unmodified
ADNI, whereas Stress Harm aggregates the degradation, missingness, and
conflict evaluations.

\subsection{In-Domain Predictive and Probabilistic Performance}
\label{sec:overall_results}

Table~\ref{tab:overall_performance} compares \textsc{TIER-MoE} with
unimodal baselines, conventional fusion
(DAFT~\cite{poelsterl2021daft}; GMU~\cite{arevalo2017gated}),
adaptive fusion (QMF~\cite{zhang2023provable}; PDF~\cite{cao2024predictive}),
and expert routing (Flex-MoE~\cite{yun2024flexmoe}; $I^2$MoE~\cite{xin2025i2moe}) across the three in-domain datasets. \textsc{TIER-MoE} achieves the best Macro-F1 and Brier score on every dataset and ranks first on 10 of the 12 dataset--metric combinations. On ADNI, it reaches a Macro-F1 of 0.648, a
3.4-percentage-point improvement over the strongest published comparator. The same advantage extends to distinct modality regimes: Macro-F1 reaches 0.884 for
image--clinical fusion on PAD-UFES-20 and 0.657 for
fundus--video fusion on FPRM. On PAD-UFES-20 and FPRM, \textsc{TIER-MoE} also obtains the strongest AUROC, Brier score, and ECE.


To isolate whether the gains arise from cross-fitted predictive evidence alone or from affinity-based expert routing alone, we compare two matched controls. OOF MetaGate receives the same cross-fitted evidence as the conditional-risk estimator but directly learns sample-specific modality weights without the expert-routing branch. \textsc{TIER-MoE} improves its Macro-F1 by 4.1\%, 5.8\%, and 5.5\% on ADNI, PAD-UFES-20, and FPRM, respectively. Affinity-MoE is an alignment-only subspace MoE that routes modality--expert pairs using expert-subspace affinity $\psi_{i,m,e}$ without the conditional modality risk. Our method further improves both Macro-F1 and Brier score over it on all three tasks. Together, these comparisons isolate the value of both components; modeling cross-fitted evidence as task-grounded modality risk rather than direct fusion weights, and combining this risk with expert compatibility rather than routing by affinity alone.

\subsection{Calibration and Discrimination of Conditional Modality Risk}
\label{sec:risk_semantics_results}

Having established the downstream prediction gains, we next test the
central quantity behind them: whether the estimated modality risk
actually predicts the loss incurred by relying on that modality.

\begin{figure}[t]
\centering
\includegraphics[width=\columnwidth]{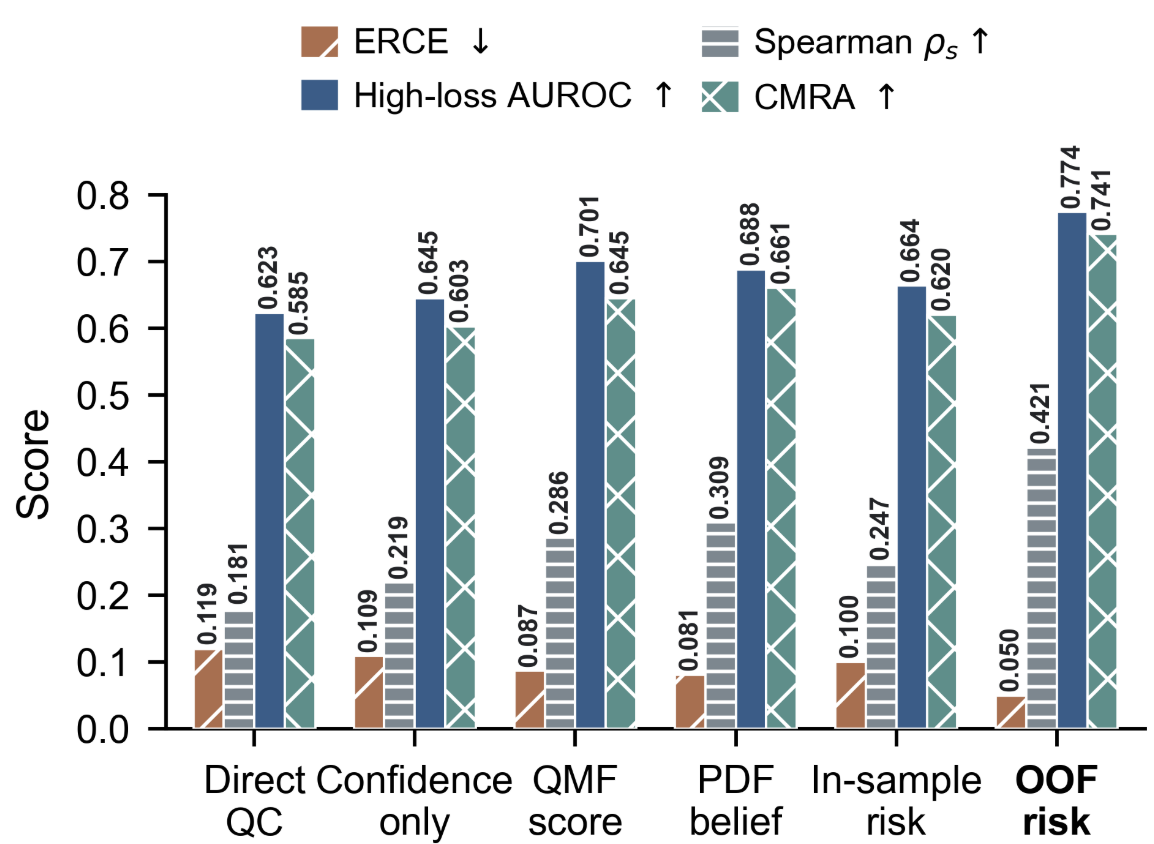}
\caption{Modality-risk semantics on held-out ADNI. ERCE and $\rho_s$
measure agreement with realized unimodal Brier loss; AUROC detects
high-loss cases; CMRA evaluates correct-modality ranking.}
\label{fig:adni_risk_semantics}
\end{figure}

Figure~\ref{fig:adni_risk_semantics} compares OOF conditional risk
with technical-quality, confidence-based, adaptive-fusion, and
in-sample-risk alternatives. OOF conditional risk achieves the best
result on all four metrics: it reduces ERCE from 0.081 for PDF to
0.050, raises Spearman correlation from 0.309 for PDF to 0.421, and
improves High-Loss AUROC from 0.701 for QMF to 0.774. These results
show that the estimated risk is both calibrated to realized loss and
discriminative of high-loss modality predictions.

CMRA tests whether the risk score ranks the correct unimodal predictor
below the incorrect one when exactly one of them is correct. OOF
conditional risk assigns lower risk to the correct modality in 74.1\%
of these cases, compared with 66.1\% for PDF. Together, the in-sample
control and the proxy-score baselines isolate the benefit of
cross-fitted loss supervision over in-sample risk fitting and
technical-quality, confidence, or fusion-derived reliability signals.


\subsection{Robustness to Unreliable Evidence}
\label{sec:robustness_results}
The preceding analysis shows that conditional risk tracks held-out unimodal loss. We next evaluate whether \textsc{TIER-MoE} remains
robust when multimodal evidence is degraded, unavailable, or
conflicting.

\begin{table}[t]
\centering
\begin{small}
\setlength{\tabcolsep}{1mm}

\begin{tabular*}{\columnwidth}{
    @{\extracolsep{\fill}}lccccc@{}
}
\toprule
\textbf{Method}
& \shortstack{\textbf{T1w Ret.}\\\textbf{AUC} $\uparrow$}
& \shortstack{\textbf{DTI Ret.}\\\textbf{AUC} $\uparrow$}
& \shortstack{\textbf{Missing}\\\textbf{F1} $\uparrow$}
& \shortstack{\textbf{Conflict}\\\textbf{F1} $\uparrow$}
& \shortstack{\textbf{Stress}\\\textbf{Harm} $\downarrow$} \\
\midrule

QMF
& $0.835$
& $0.862$
& $0.494$
& $0.516$
& $0.171$ \\

PDF
& $\underline{0.851}$
& $\underline{0.875}$
& $0.501$
& $\underline{0.539}$
& $\underline{0.158}$ \\

Flex-MoE
& $0.848$
& $0.860$
& $\underline{0.523}$
& $0.527$
& $0.162$\\

Affinity-MoE
& $0.812$
& $0.847$
& $0.487$
& $0.508$
& $0.184$ \\

\midrule

\textbf{TIER-MoE}
& $\mathbf{0.902}$
& $\mathbf{0.921}$
& $\mathbf{0.531}$
& $\mathbf{0.574}$
& $\mathbf{0.109}$ \\

\bottomrule
\end{tabular*}
\end{small}

\caption{Robustness on ADNI under progressive modality degradation,
single-modality missingness, and natural unimodal disagreement.
Retention AUC integrates clean-normalized Macro-F1 over degradation
severity.}
\label{tab:robustness}

\end{table}

\textbf{\textsc{TIER-MoE} preserves performance under modality degradation.}
Table~\ref{tab:robustness} shows that \textsc{TIER-MoE} achieves
retention AUCs of 0.902 and 0.921 under progressive T1w and DTI
degradation, exceeding PDF, the strongest baseline by 5.1 and
4.6 percentage points, respectively. 
The advantage holds when either modality is degraded, including the globally stronger T1w modality, indicating that the model does not depend on a fixed modality preference.

\textbf{\textsc{TIER-MoE} remains robust to missingness and unimodal disagreement.}
\textsc{TIER-MoE} attains the highest Missing F1 of 0.531 and Conflict
F1 of 0.574. These metrics probe complementary failure modes:
Missing F1 measures prediction using the remaining modality when one
source is unavailable, whereas Conflict F1 measures fusion performance
when the two unimodal predictors produce competing decisions. The
strong results in both cases indicate that \textsc{TIER-MoE} can
operate effectively with incomplete evidence while resolving
cross-modal disagreement without consistently favoring one modality.

\textsc{TIER-MoE} further reduces Stress Harm from 0.158 for PDF to
0.109, a 31.0\% relative reduction. Together with the retention
results, these findings show that \textsc{TIER-MoE} degrades more
gracefully and is less likely to override correct unimodal evidence
under modality degradation, missingness, and disagreement.

\subsection{Zero-Shot ADNI-to-OASIS-3 Transfer}
\label{sec:external_results}

We next examine whether the learned reliability and routing mechanisms
transfer to an external cohort without adaptation. Under the frozen
ADNI-to-OASIS-3 protocol, Table~\ref{tab:oasis_external} shows that
\textsc{TIER-MoE} achieves the best balanced accuracy (0.568),
Macro-F1 (0.535), Brier score (0.249), and ECE (0.078). Its AUROC
of 0.776 is within 0.004 of the best result. The strongest external
results therefore occur in class-balanced prediction and probability
quality, while ranking performance remains competitive.

The transfer extends beyond the final prediction to the learned risk
semantics. Without OASIS-3 training or recalibration, OOF conditional
risk reduces ERCE from 0.101 for PDF to 0.074, raises Spearman
correlation from 0.261 for PDF to 0.347, improves High-Loss AUROC
from 0.671 for QMF to 0.724, and increases CMRA from 0.626 for PDF
to 0.684. Thus, the estimated risks retain their calibration to
realized modality loss, high-loss discrimination, and correct-modality
ordering under cohort shift. The zero-shot transfer therefore extends
beyond downstream classification to the task-grounded reliability
semantics central to \textsc{TIER-MoE}.

\subsection{Ablation Studies}
\label{sec:ablation_results}


Finally, we use controlled ablations to examine three design questions:
whether the reliability and interaction branches provide complementary
capabilities; whether conditional modality risk improves affinity-based expert routing; and whether OOF supervision is necessary for task-grounded
risk estimation. The first three variants evaluate frozen branches or pathways from the same trained model, whereas the remaining variants modify the routing criterion or risk supervision.


\textbf{The prediction branches exhibit complementary strengths.}
Table~\ref{tab:main_ablation} shows that the interaction branch
achieves higher Macro-F1 than the reliability anchor
(0.621 vs.\ 0.603) and a lower Brier score
(0.204 vs.\ 0.227), whereas the reliability anchor yields lower
Clean Harm (0.126 vs.\ 0.153). Within the interaction branch,
removing the sparse expert residuals reduces Macro-F1 from 0.621 to
0.615 and worsens Brier score from 0.204 to 0.211, showing that the
routed experts add predictive value beyond the shared path. The shared
path alone produces slightly lower Clean Harm than the full interaction
branch (0.148 vs.\ 0.153), indicating a trade-off between specialized
interaction and decision stability. Combining the reliability and
interaction predictions with the validation-selected $\gamma$ achieves
the best Macro-F1 (0.648), Brier score (0.189), and Clean Harm
(0.091).

\textbf{Risk-aware routing and OOF supervision both contribute to the
full model.}
Removing conditional risk from the routing score by setting
$\eta=0$ reduces Macro-F1 from 0.648 to 0.633, worsens Brier score
from 0.189 to 0.199, and increases Clean Harm from 0.091 to 0.114.
Separately, replacing the OOF-derived risk targets with in-sample
supervision yields a Macro-F1 of 0.629, a Brier score of 0.205, and
a Clean Harm of 0.121. These comparisons isolate the contributions
of risk-conditioned expert routing and label-honest OOF supervision
to predictive performance, probability quality, and harmful-fusion
control.




\begin{table}[t!]
\centering
\begin{small}
\setlength{\tabcolsep}{1mm}
\begin{tabular*}{\columnwidth}{
    @{\extracolsep{\fill}}lccccc@{}
}
\toprule
\textbf{Model}
& \textbf{AUROC} $\uparrow$
& \textbf{BA} $\uparrow$
& \textbf{F1} $\uparrow$
& \textbf{Brier} $\downarrow$
& \textbf{ECE} $\downarrow$ \\
\midrule

T1w only
& $0.706$
& $0.463$
& $0.435$
& $0.325$
& $0.149$ \\

DTI only
& $0.674$
& $0.429$
& $0.398$
& $0.346$
& $0.168$ \\
\midrule

I$^{2}$MoE
& $0.744$
& $0.509$
& $0.483$
& $0.287$
& $0.111$ \\

\midrule
QMF
& $0.753$
& $0.519$
& $0.493$
& $\underline{0.256}$
& $\underline{0.086}$ \\

PDF
& $\mathbf{0.780}$
& $0.512$
& $0.477$
& $0.257$
& $0.092$ \\

\midrule
OOF MetaGate
& $0.748$
& $0.476$
& $0.447$
& $0.286$
& $0.119$ \\

Affinity-MoE
& $0.773$
& $\underline{0.541}$
& $\underline{0.508}$
& $0.291$
& $0.104$ \\

\midrule
\textbf{TIER-MoE}
& $\underline{0.776}$
& $\mathbf{0.568}$
& $\mathbf{0.535}$
& $\mathbf{0.249}$
& $\mathbf{0.078}$ \\

\bottomrule
\end{tabular*}
\end{small}

\caption{Frozen ADNI-to-OASIS-3 cross-cohort evaluation.
All model hyperparameters are selected using ADNI
only and applied to OASIS-3 without adaptation.}
\label{tab:oasis_external}
\end{table}


\begin{table}[t]
\centering
\begin{small}
\setlength{\tabcolsep}{1mm}

\begin{tabular*}{\columnwidth}{
    @{\extracolsep{\fill}}lccc@{}
}
\toprule
\textbf{Variant}
& \shortstack{\textbf{Macro-F1}$\uparrow$}
& \shortstack{\textbf{Brier}$\downarrow$}
& \shortstack{\textbf{Clean}\\\textbf{Harm}$\downarrow$} \\
\midrule

Reliability anchor only ($\gamma=0$)
& $0.603$
& $0.227$
& $0.126$ \\

Interaction branch only ($\gamma=1$)
& $0.621$
& $0.204$
& $0.153$ \\

Shared-path only ($\gamma=1$, $h_i^{\mathrm{sparse}}$ =0)
& $0.615$
& $0.211$
& $0.148$

\\
\shortstack[l]{Affinity-only routing 
($\eta=0$)}
& $\underline{0.633}$
& $\underline{0.199}$
& $\underline{0.114}$ \\

\shortstack[l]{In-sample (Non-OOF) risk}
& $0.629$
& $0.205$
& $0.121$ \\

\midrule

\textbf{Full TIER-MoE}
& $\mathbf{0.648}$
& $\mathbf{0.189}$
& $\mathbf{0.091}$ \\

\bottomrule
\end{tabular*}
\end{small}

\caption{Component and routing ablations on ADNI. The first three rows
evaluate frozen branches or pathways from the same trained model;
$\eta=0$ removes conditional risk from the routing score, and the
in-sample variant replaces OOF-derived risk targets.}
\label{tab:main_ablation}
\end{table}

\section{Conclusion}

We introduce \textsc{TIER-MoE}, which learns subject-specific modality
risk from out-of-fold losses, combines it with expert-subspace affinity
for sparse routing, and preserves cross-modal interaction through a
shared path. Across three in-domain tasks, \textsc{TIER-MoE} achieves
the best Macro-F1 and Brier score; its risks track realized unimodal
loss, and it remains robust to degradation, missingness, and
disagreement, with strong zero-shot transfer to OASIS-3. These results
show that modality reliability, cross-modal complementarity, and expert
compatibility are distinct yet coordinated factors in multimodal fusion.

\bibliography{references}

\end{document}